\documentclass[10pt,twocolumn,letterpaper]{article}
\usepackage{comment}
\usepackage{cvpr}              

\usepackage[accsupp]{axessibility}
\usepackage{graphicx}
\usepackage{amsmath}
\usepackage{amssymb}
\usepackage{booktabs}

\usepackage{caption,subcaption}
\usepackage[linesnumbered,ruled,vlined]{algorithm2e}
\SetKwInput{KwInput}{Input}
\SetKwInput{KwOutput}{Output}
\usepackage{multirow}
\usepackage[pagebackref,breaklinks,colorlinks]{hyperref}
\usepackage{animate}
\usepackage{tabularx}
\usepackage[capitalize]{cleveref}
\crefname{section}{Sec.}{Secs.}
\Crefname{section}{Section}{Sections}
\Crefname{table}{Table}{Tables}
\crefname{table}{Tab.}{Tabs.}

\def\confName{CVPR}
\def\confYear{2022}

\begin{document}

\title{Video alignment using unsupervised learning of local and global features}

\author{
    Niloufar Fakhfour$^{1}$,    Mohammad ShahverdiKondori$^{1}$,    Sajjad Hashembeiki$^{1}$,   Mohammadjavad Norouzi$^{1}$, \\ \\
    Hoda Mohammadzade$^{1}$ \\  \\
    $^1$Sharif University of Technology \\ \\ 
    {\tt\small {\{niloufar.fakhfour, m.shah, s.hashem, javad.noroozi99, hoda\}}@sharif.edu} 
}

\twocolumn[{%
\renewcommand\twocolumn[1][]{#1}%
\maketitle
}]

\begin{abstract}
In this paper, we tackle the problem of video alignment, the process of matching the frames of a pair of videos containing similar actions. The main challenge in video alignment is that accurate correspondence should be established despite the differences in the execution processes and appearances between the two videos. We introduce an unsupervised method for alignment that uses global and local features of the frames. In particular, we introduce effective features for each video frame by means of three machine vision tools: person detection, pose estimation, and VGG network. Then the features are processed and combined to construct a multidimensional time series that represent the video. The resulting time series are used to align videos of the same actions using a novel version of dynamic time warping named Diagonalized Dynamic Time Warping(DDTW). The main advantage of our approach is that no training is required, which makes it applicable for any new type of action without any need to collect training samples for it. Additionally, our approach can be used for framewise labeling of action phases in a dataset with only a few labeled videos. For evaluation, we considered video synchronization and phase classification tasks on the Penn action \cite{penn} and subset of UCF101 \cite{UCF} datasets. Also, for an effective evaluation of the video synchronization task, we present a new metric called Enclosed Area Error(EAE). The results show that our method outperforms previous state-of-the-art methods, such as TCC \cite{TCC}, and other self-supervised and weakly supervised methods. 
\end{abstract}

\section{Introduction}
Many sequential processes happen daily in the world. Waking up, drinking water, and growing a plant are examples of sequential processes that are always happening. Although these processes are performed with different varieties and qualities, all the processes that show a specific action have common time points. For example, drinking a glass of water may happen at different speeds, places, and containers, but all the processes that indicate drinking water consist of 3 main steps: lifting the glass, drinking water, and lowering the glass. As a result, each process, or in other words, each action, consists of one or more phases, which are the same in terms of the order of occurrence in all similar processes.
Video alignment is a method in which the frames of the videos of two identical actions that differ in things such as scene, camera angle, and speed are matched to each other.

\begin{figure}[h] 
  \centering
  \includegraphics[width=8cm, height=6cm]{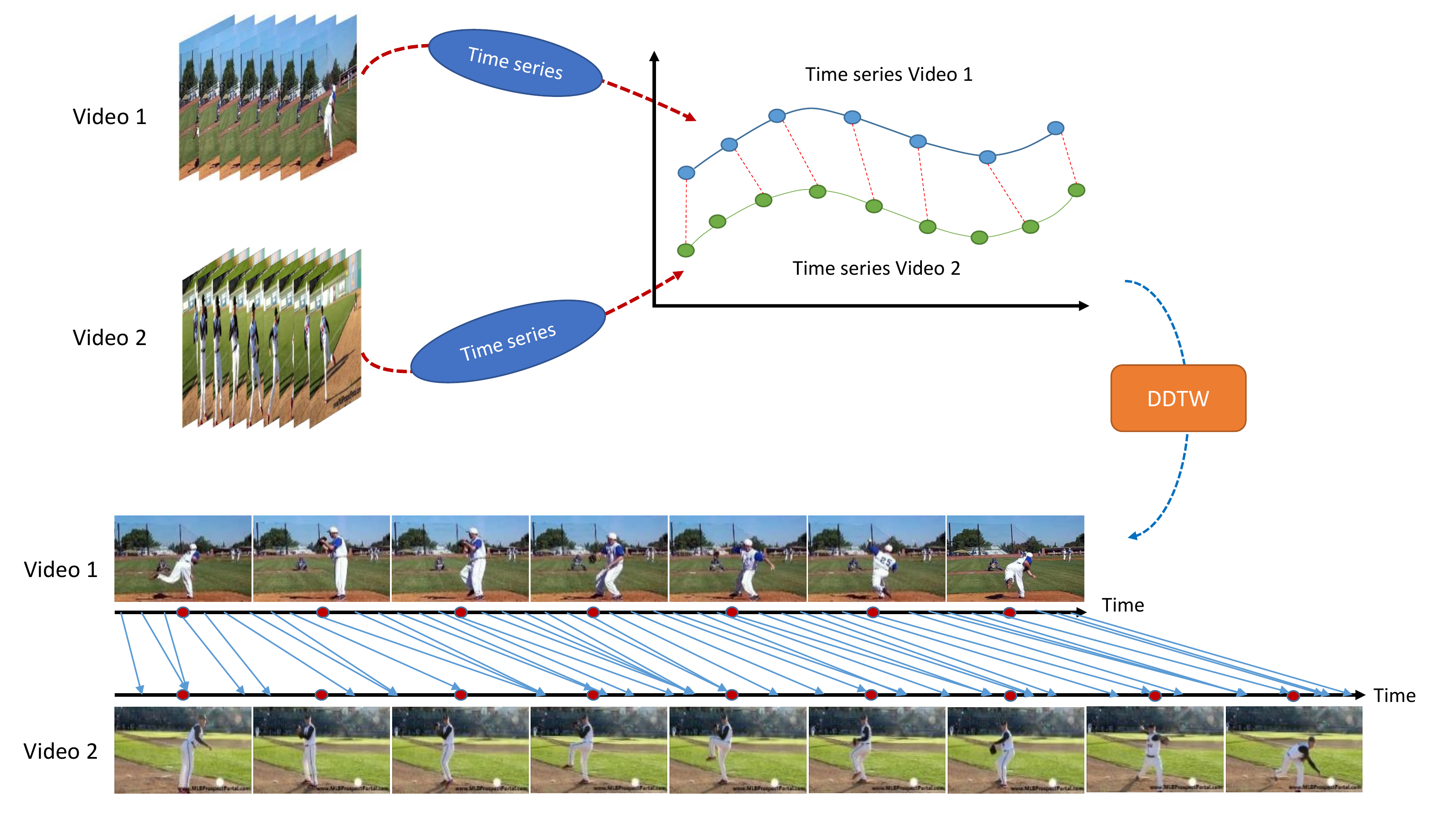}
  \caption{We propose an unsupervised method to align pairs of videos that present the same actions. We model a video as a time series which consists of global and local features extracted from each frame. In addition,  we introduce a novel DTW, called Diagonalized Dynamic Time Warping (DDTW), to find corresponding frames in each pair of videos.}
  \label{header}
\end{figure}

The main challenge in the task of video alignment is the difference in the execution process and the appearance of the frames in videos containing the same action. For example, an action such as picking up an object from the ground can be done in various ways while recorded by the camera. This action may be done once in $5$ seconds and another time in $10$ seconds. The object that is risen from the ground may be a ball or a cup, big or small, red or blue. Also, the camera may record this action from the front, side, or any other angle. All these differences cause video alignment and choosing the correct method for video modeling to face many challenges.

In recent years, much research has been done in action recognition, anomaly detection, tracking, etc. However, video alignment has received less attention, while video alignment can be used to improve all of the above. For example, in \cite{normalized_pose}, novel normalized pose features invariant to video subjects’ anthropometric characteristics are introduced. The method was evaluated in the task of action recognition, and significant results were achieved. In \cite{few-shot-action-recognition}, the action recognition problem is considered as two separate problems, action duration misalignment, and action evolution misalignment. Based on this assumption, a two-stage action alignment network is presented in their work. Video Classification \cite{few-shot-classification , few-shot-classiciation2} and action detection \cite{action-detection, action-detection2, action-detection3} are other examples of the applications of video alignment that have received attention in recent years.

In the field of video alignment, self-supervised \cite{TCC , aligning_in_space_and_time , learning-by-aligning-videos-in-time , learning-to-align-sequential-actions , ctw} and weakly-supervised methods \cite{D3TW , representation-learning-via-global-temporal-alignment, weakly3, weakly4} have been presented in recent years. In some works, video alignment has been tried to solve using Dynamic Time Warping(DTW) \cite{few-shot-classiciation2 , D3TW , ctw}. Since DTW is not derivable and cannot be implemented using neural networks, in these works, some modified types of DTW, such as soft-dtw, which are derivable, have been used \cite{few-shot-classiciation2 , D3TW }. Another category of video alignment methods is based on cycle consistency loss \cite{TCC , aligning_in_space_and_time , representation-learning-via-global-temporal-alignment, cycle-consistency}. In \cite{TCC}, they presented a self-supervised method for learning correspondences between frames in the time domain. In this article, the network is trained based on the cycle-consistency cost function, and then the trained network is used to match the frames of a pair of videos with each other. Unlike \cite{TCC}, \cite{aligning_in_space_and_time} deals with correspondence learning in both time and space domains based on cross-cycle stability. In \cite{learning-by-aligning-videos-in-time , representation-learning-via-global-temporal-alignment}, the network is trained based on frame level and video level simultaneously. In \cite{learning-by-aligning-videos-in-time}, the network is trained based on a cost function including two terms, soft-dtw and temporal regularization. In \cite{representation-learning-via-global-temporal-alignment}, a weakly supervised method is presented based on a cost function consisting of dtw and cycle-consistency. In \cite{learning-to-align-sequential-actions}, another look at the subject of video alignment is given. In this work, video representation is learned to align two videos while the possibility of background frames, redundant frames, and non-monotonic frames are considered.

One of the shortcomings of the existing methods is the need to train deep networks for each class of action, which requires a lot of training samples from each action. In this work, we present an unsupervised method that can be used to align pairs of videos containing any action without any need to train a network.

The main contribution of this article is, using an unsupervised approach for representing a video as a multidimensional time series representing features of its frames over time. To construct these features, we simultaneously employ box and pose features to extract local details, while using the VGG network to capture global features.

This combination of local and global features provides an effective representation that models the details of each phase of an action, enabling accurate video alignment. These features can be used individually for phase classification tasks or can be aligned using Dynamic Time Warping (DTW) to synchronize the time series of videos. As shown in Figure \ref{header}, the time series of videos are aligned using a modified version of DTW, known as DDTW, which is introduced in this study.

To evaluate the effectiveness of the proposed features, we compared their performance with existing self-supervised \cite{SAL,TCN,TCC} and weakly supervised \cite{representation-learning-via-global-temporal-alignment} methods in phase classification tasks on the PennAction \cite{penn} and a subset of the UCF101 \cite{UCF}, known as UCF9, datasets. Additionally, a new evaluation metric is introduced in this work to compare the performance of different alignment methods.

In summary, our contributions include the following:
\begin{itemize}
    \item Presenting an unsupervised method to align two videos.
    \item Modeling videos using their global and local features as well as their static and dynamic features.
    \item Presenting a modified DTW method for aligning time series with limited deviation.
    \item Presenting a new metric to compare the performance of alignment methods.
\end{itemize}


\section{Method}

\begin{figure*}[h] 
  \centering
  \includegraphics[width=\linewidth, height=8cm]{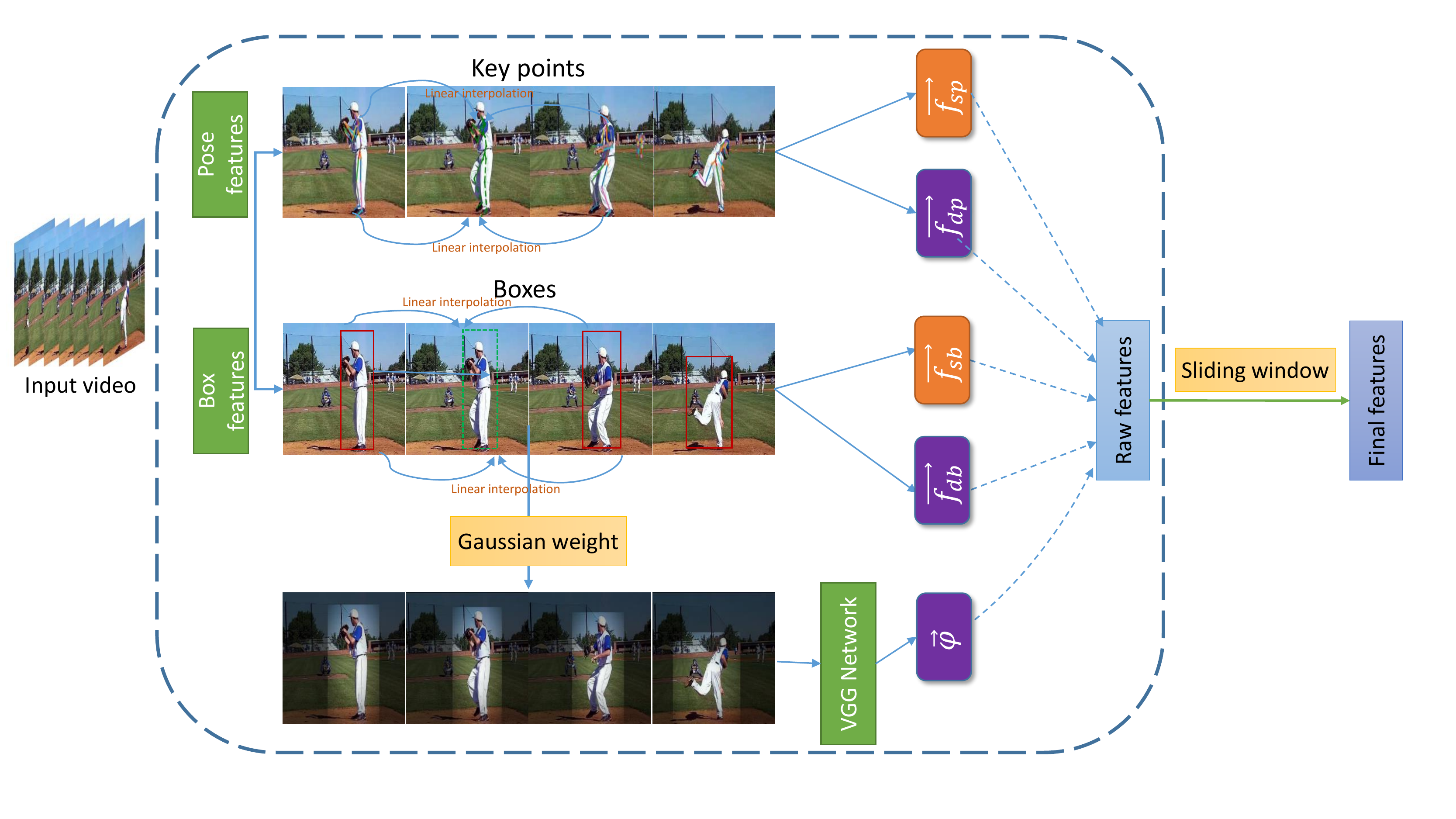}
  \caption{In our method, two types of features are used to build time series: local features, including (pose and box features) and global features. Depending on the extracted pose and box, static and dynamic features are calculated for each image. To calculate the global features, we multiply the pixels of each frame by Gaussian weight according to the extracted box and apply the final frame to the input of the VGG network and extract the global features based on it.}
  \label{algo}
\end{figure*}

This section introduces our unsupervised method for aligning videos containing similar actions. The core idea is to model a video as a time series, representing both global and local features extracted from each frame. Global features, which encapsulate comprehensive information about the entire frame, are derived using the VGG pre-trained network. Local features, focusing on the subject's movements, are obtained through pose estimation and person detection algorithms. By combining these global and local features, our method effectively captures the essential details required for distinguishing between different phases of an action and for synchronizing videos of the same action. Figure \ref{algo} provides an illustrative overview of our proposed video alignment method.

\subsection{Feature Extraction}
Features play a vital role in the areas of image processing\cite{feature_extraction, feature-extraction2, feature-extraction3} and machine vision, as extracting more valuable features leads to a better result. We use two kinds of features to model a video as a time series: global features, which are used to model the entire frame, and local features, which are used to model the details of the main subject performing the action. Features are extracted using three methods: pose estimation, person detection, and VGG network.

\subsubsection{Local Features} 
Each action is performed by one or more main subjects. Although similar actions might differ in the scene, speed, and recording quality, the main subjects follow the same process. One important step in characterizing different actions is to represent the details of the movements of the main subject by a number of features. We call these features local features. Local features consist of two types: static and dynamic, which are responsible for representing the within-frame and between-frame information related to the main subject, respectively.

\paragraph{Static Features}
In this work, static features refer to the features extracted from each frame independently from other frames. Static features consist of static pose features and static box features. These features represent the details of the current location of the main subject.

\subparagraph{Static Box Features}
Static box features model the main subject's global motion and location. To extract these features, we track the main subject using the deep sort algorithm\cite{deep_sort} with Yolo V5 \cite{yolov5deepsort} in order to extract the box of the main subject in each frame. After extracting the subject's box in each frame, the length-to-width ratio and center of the boxes are used as static box features. These features explain the change in position and angle of the main subject's body in each frame. In order to remove the effect of the initial position and the appearance characteristics of the main subject, the center of the box in the first frame is placed at the coordinate center, and the height-to-width ratio of the box is set to $1$. These features in the other frames are also normalized according to the changes in the first frame:

\begin{equation}
f_{sb}^{c}(n) = c(n) - c(1) \quad \quad \forall n
\end{equation}

\begin{equation}
f_{sb}^{r}(n) = \frac{r(n)}{r(1)} \quad \quad \forall n
\end{equation}

where $c(n)$ and $r(n)$ denote the coordinates of the center and the height-to-width ratio of the subject's box in frame $n$, respectively. Also, $f_{sb}^{m}(n)$ denotes the static box feature ($m = c$ and $m = r$ refer to the center and the height-to-width ratio of the box, respectively). Therefore, three features are extracted from each frame.

\subparagraph{Static Pose Features}

Static pose features consist of the positions of the key points of the main subject in each frame. 

Human pose estimation refers to determining the position of human joints (known as body key points) \cite{pose_survey, pose, pose-estimation-analysis}. Key points extracted from pose estimation contain helpful information and details about the gestures and actions of the subject. 

In this work, to extract the key points of the main subject, we use MeTRAbs \cite{human_pose} algorithm for pose estimation, which extracts 24 key points. After the key points are extracted, to remove the effect of the initial position of the main subject in the first frame, we shift the hip joint in the first frame to the coordinate center. We also shift all key points in all frames by the same translation vector obtained in the first frame.

\begin{equation}
f_{sp}^{m}(n) = k^{m}(n) - k^{1}(1) \quad \quad \forall m,n
\end{equation}

where $k^{m}(n)$ denotes the 2D coordinates of the $m$-th key point in frame $n$ ($m = 1$ refers to the hip joint key point) and $f^{m}_{sp}(n)$ denotes the static pose feature corresponding to the $m$-th key point in frame $n$.
Finally, $48$ static pose features are extracted from each frame.

\paragraph{Dynamic Features}
In addition to static features, to appropriately model a video of an action, some features for representing the changes between frames are also required. In this work, dynamic features refer to the features extracted based on the changes between successive frames. More specifically, these features consist of displacement vectors between the static features.  

\subparagraph{Dynamic Box Features}
The displacement vector of the center position and the changes in the height-to-width ratio of the box constitute the first part of the dynamic features, which can effectively model the progression of an action. This part of dynamic features consists of the displacement vector between the static box feature in each frame and its previous frame as:

\begin{equation}
f_{db}^{m}(n) = f_{sb}^{m}(n) - f_{sb}^{m}(n-1) \quad \quad \forall m,n
\end{equation}

where $f_{db}^{m}(n)$ denotes the dynamic pose feature in frame $n$. Note that $f^{m}_{db}(1)$ is considered to be zero. Finally, three dynamic box features are extracted for each frame.

\subparagraph{Dynamic Pose Features}
The second part of dynamic features consists of the displacement vector between the key points in each frame and its previous frame as:

\begin{equation}
f_{dp}^{m}(n) = k^{m}(n) - k^{m}(n-1) \quad \quad \forall m,n
\end{equation}

where $f^{m}_{dp}(n)$ denotes the dynamic pose feature for the $m$-th key point in frame $n$. Note that $f^{m}_{dp}(1)$ is considered to be zero. Finally, $48$ dynamic pose features are extracted for each frame.

\paragraph{Interpolation for Missing Data}
Each of the pose and detection algorithms may fail to extract the key points and boxes in some frames. Linear interpolation is used to estimate the missing key points and boxes using those in the most recent frames that are before and after the current frame.

\subsubsection{Global Features} 
Our goal is to use a combination of local and global features of the frames over time to represent videos. Global features represent information over the whole frame. Obviously, an action is performed by a subject, and therefore local features are more directly related to the type of action being performed than global features. However, the objects around the main subject, the appearance of the subject, and the background serve as important side information to represent an action.

\begin{figure*}[h] 
  \centering
  \includegraphics[width=\linewidth]{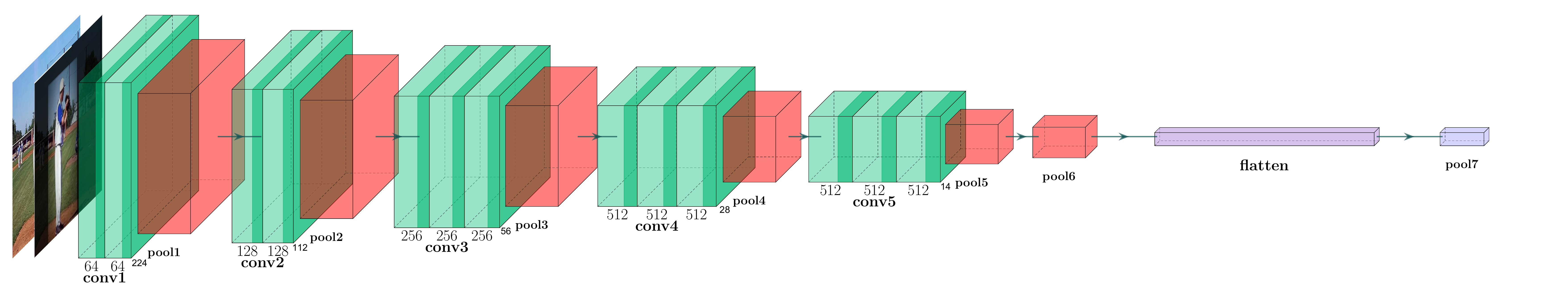}
  \caption{VGG network is used to calculate the global features. The input of the network is a weighted frame based on the truncated 2D Gaussian weight. In order to adapt the network, the last three fully connected layers are replaced with the 2D max-pooling layer of stride $(1, 1)$ and filter size $(7, 7)$, followed by the flatten layer, and then the 1D max-pooling layer with a stride of $1$ and a size of $8$.} 
  \label{vgg}
\end{figure*}

We use VGG16 network \cite{VGG}, which is pre-trained on the Imagenet dataset \cite{imagenet} to extract global features. To adapt the network, we replace the fully connected layers with the 2D max-pooling layer of stride $(1, 1)$ and filter size $(7, 7)$, flatten layer, and the 1D max-pooling layer of stride $1$ and size $8$. In order to focus more on the subject than other details of the scene, a truncated 2D Gaussian weight mask is applied to the pixels of the input frame before feeding it to the network. Figure \ref{vgg} illustrates the final network.
The truncated 2D Gaussian weight mask is designed according to the following points:

\begin{itemize}
    \item Pixels located inside the subject box, with a high probability, are more related to the action.
    \item A margin is considered for the box boundaries to reduce the error caused by the box extraction algorithm as well as not to attenuate the objects that are very close to the subject. More specifically, the height and width of the box are increased by $20$ pixels\footnote{Different values, including those proportional to the height of the box and other fixed values, were examined. among these options, it was found that a value of $20$ pixels yielded the best result}.
\end{itemize}

The weight of the mask is constant outside of the adjusted box and is $0.2$ less than the smallest weight of the 2D Gaussian on the boundaries of the box.

\begin{equation}
g_{x,y} = \exp{(- \frac{(x-x_{center})^{2} + (y-y_{center})^{2}}{2})}
\end{equation}

\begin{equation}
w_{x,y} = \begin{cases}
                    g_{x,y}  &  p_{x,y}\in mbox \\
                    g_{min} - 0.2 &  p_{x,y}\notin mbox
                \end{cases}
\end{equation}

$w_{x,y}$ denotes the weight that should be multiplied by pixel $p(x,y)$. $mbox$ indicates the box with the margin. $g_{min}$ represents the lowest coefficient on the boundary of the $mbox$. Finally, $64$ features are extracted for each frame as global features.

\subsection{Construction of Time Series}
After extracting the local and global feature vectors, they are concatenated, and a feature vector with a length of $166$ is constructed for each frame. After that, the feature vectors of the frames of each video form a multidimensional time series that represents the video. Note that at each time step, the resulting time series has a $166$ dimensional feature vector. In order to reduce the noise of the extracted features, a moving vector average with a window length equal to $5$ is used. Also, for each time series, the mean and variance of each element of the feature vector are normalized over the time series. The final method of constructing a time series from a video is shown in Figure \ref{algo}.

\subsection{DDTW} 
\begin{figure}[h] 
    \centering 
    \includegraphics[width=\linewidth]{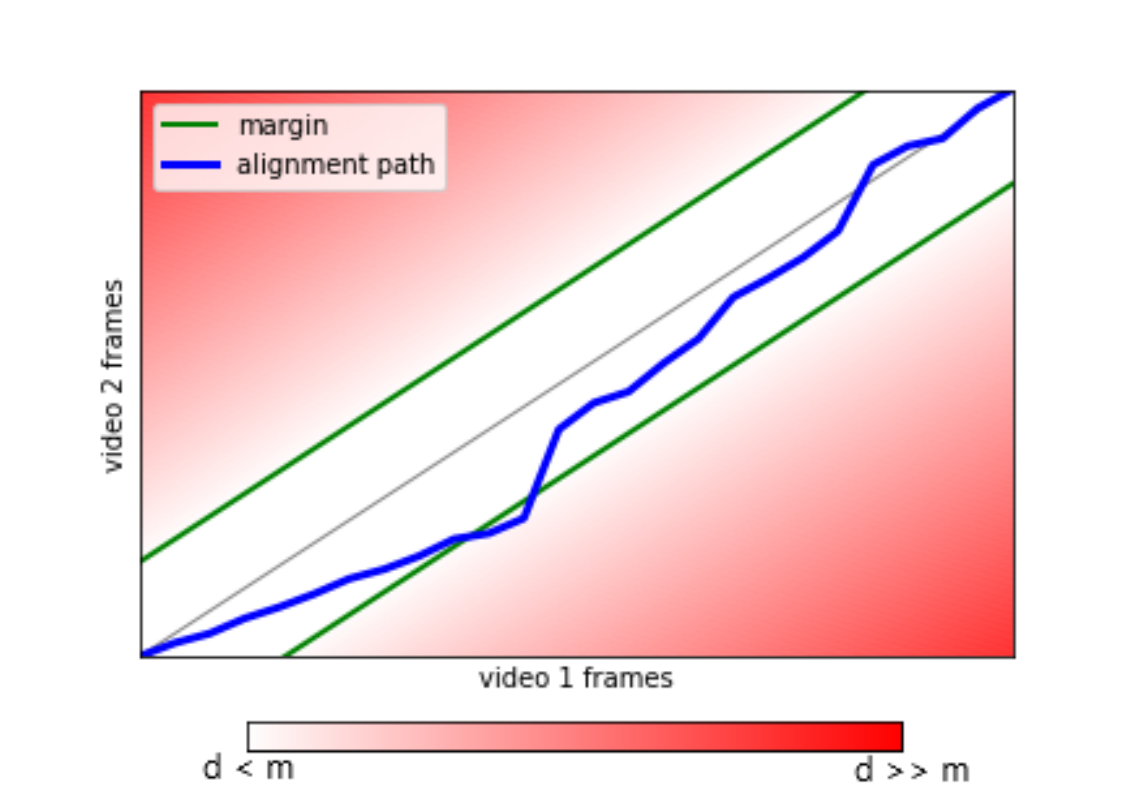}
\caption{DDTW method, the green lines parallel to the diagonal show the margin. The blue path shows the alignment of frames, and going out of the margin results in a penalty, which is calculated according to the distance from the diagonal.} 
\label{ddtw}
\end{figure}

Dynamic time warping(DTW) \cite{DTW, DTW-2} is one of the most popular algorithms for measuring the similarity between a pair of sequences and computing the best way to align them, no matter whether their lengths are equal or not. Different kinds of DTW have been developed in various fields \cite{DGTW, DTW-app-speech, Soft-DTW}, and also some works have used DTW in video alignment tasks \cite{DTW-alignment, dtw-in-alignment}. In this work, a novel method called Diagonalized Dynamic Time Warping(DDTW) is introduced, which is a generalization of the DTW method. Consider the sets $X = \{x_1, x_2, \cdots , x_n\}$ and $Y = \{y_1, y_2, \cdots, y_k\}$ as the frames of the first and second video, respectively, and build an $k \times n$ table $D$ such that $D_{i,j}$ is the Euclidean distance between the feature vector of $x_i$ and $y_j$.  In conventional DTW, the algorithm finds the best alignment (a path from the down-left corner of the table to the top-right corner which can only move in three directions $\rightarrow  \uparrow \nearrow $ ) with the minimum sum of $D_{i, j}$'s. In DDTW, a penalty coefficient is considered if the path gets further than a threshold from the diagonal. The reason for this penalty is the observation that the frames of similar actions performed by different subjects are almost linearly corresponding to each other. Therefore, the alignment path is close to the diagonal. We consider a margin $m$ and build a new table $D^\prime$ as: 

\begin{equation}
D^\prime_{i,j} = \begin{cases}
                    D_{i,j}  &  d \leq m\\
                    D_{i,j} (1 + \lambda (d - m)) & d > m
                \end{cases}
\end{equation}

where $\lambda$ is the DDTW coefficient and $d$ is the orthogonal Euclidean distance between the table's $(i,j)$- cell and the diagonal. The distance $d$ can be calculated using the formula for the distance of a point from a line in the plane; in our case, the final formula for $d$ is the following: 
\begin{equation} 
d = \frac{|\frac{k}{n} \times i - j|}{\sqrt{\frac{k^2}{n^2} + 1}}
\end{equation}
In the end, the best path, which has the minimum sum of the $D^\prime_{i,j}$'s should be found. (Figure \ref{ddtw}) 

\begin{figure*}[t] 
\centering
\begin{subfigure}[H]{.33\linewidth} 
  \centering 
  \includegraphics[width=\linewidth]{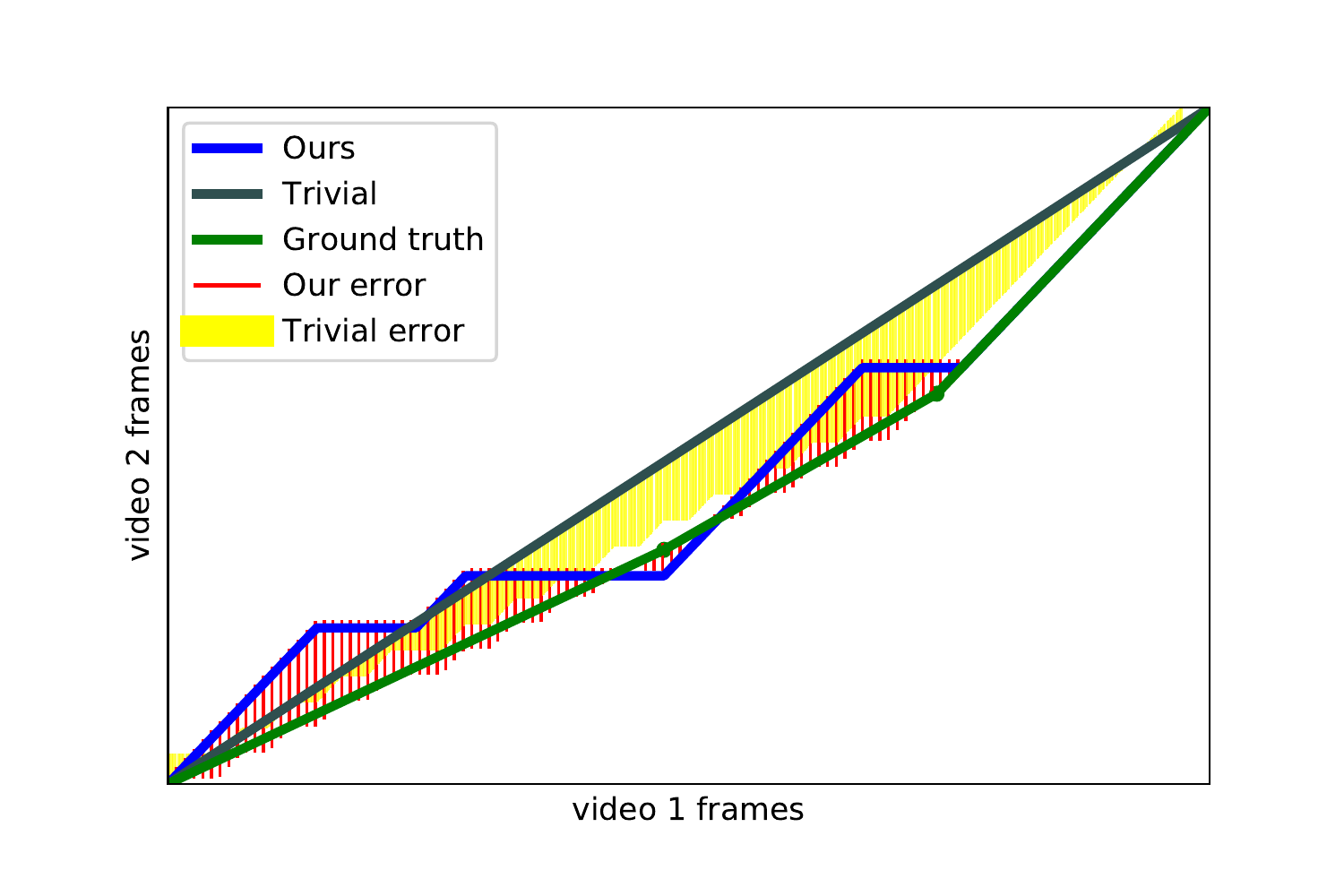}  
\end{subfigure}
\begin{subfigure}[H]{.33\linewidth}
  \centering
  \includegraphics[width=\linewidth]{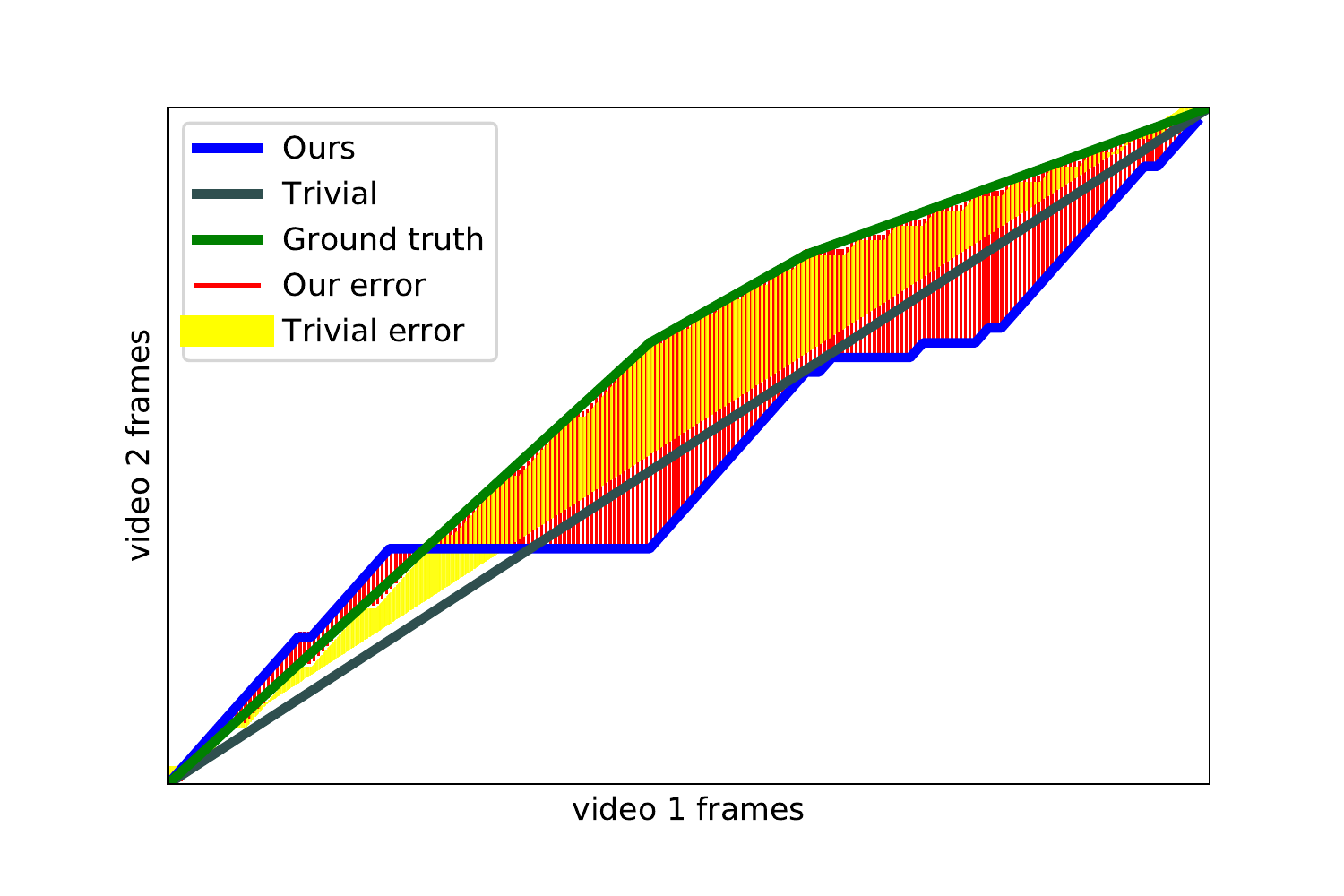}  
\end{subfigure}
\begin{subfigure}[H]{.33\linewidth}
  \centering
  \includegraphics[width=\linewidth]{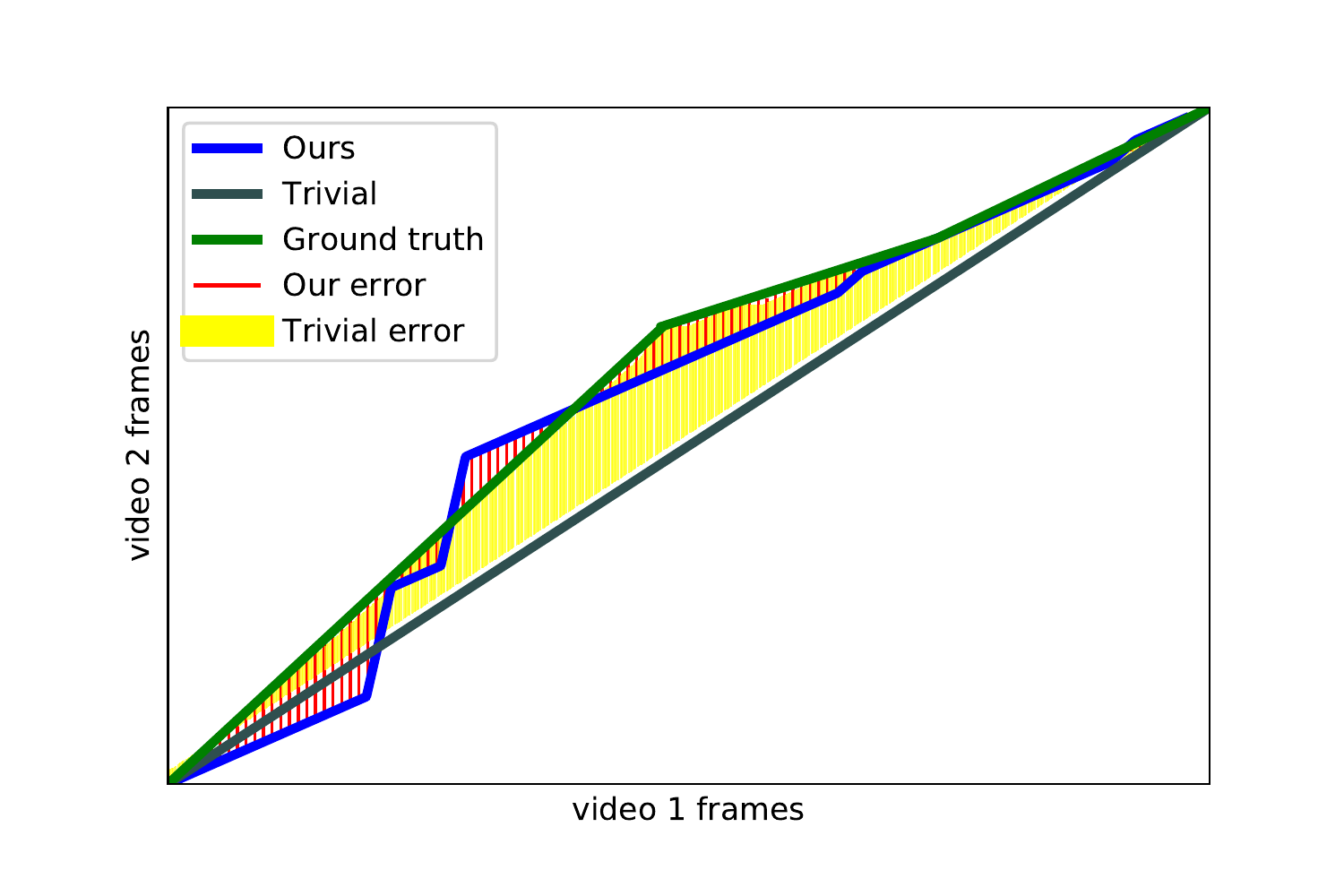}  
\end{subfigure}
\caption{The enclosed area for our predicted and trivial path for three pairs of videos. For the trivial method, the alignment path is the straight line passing through the lower left and upper right corners of the table.}
\label{trivial-eae}
\end{figure*}

\subsection{Enclosed Area Error(EAE)}
\begin{figure}[h] 
  \centering 
  \includegraphics[width=\linewidth]{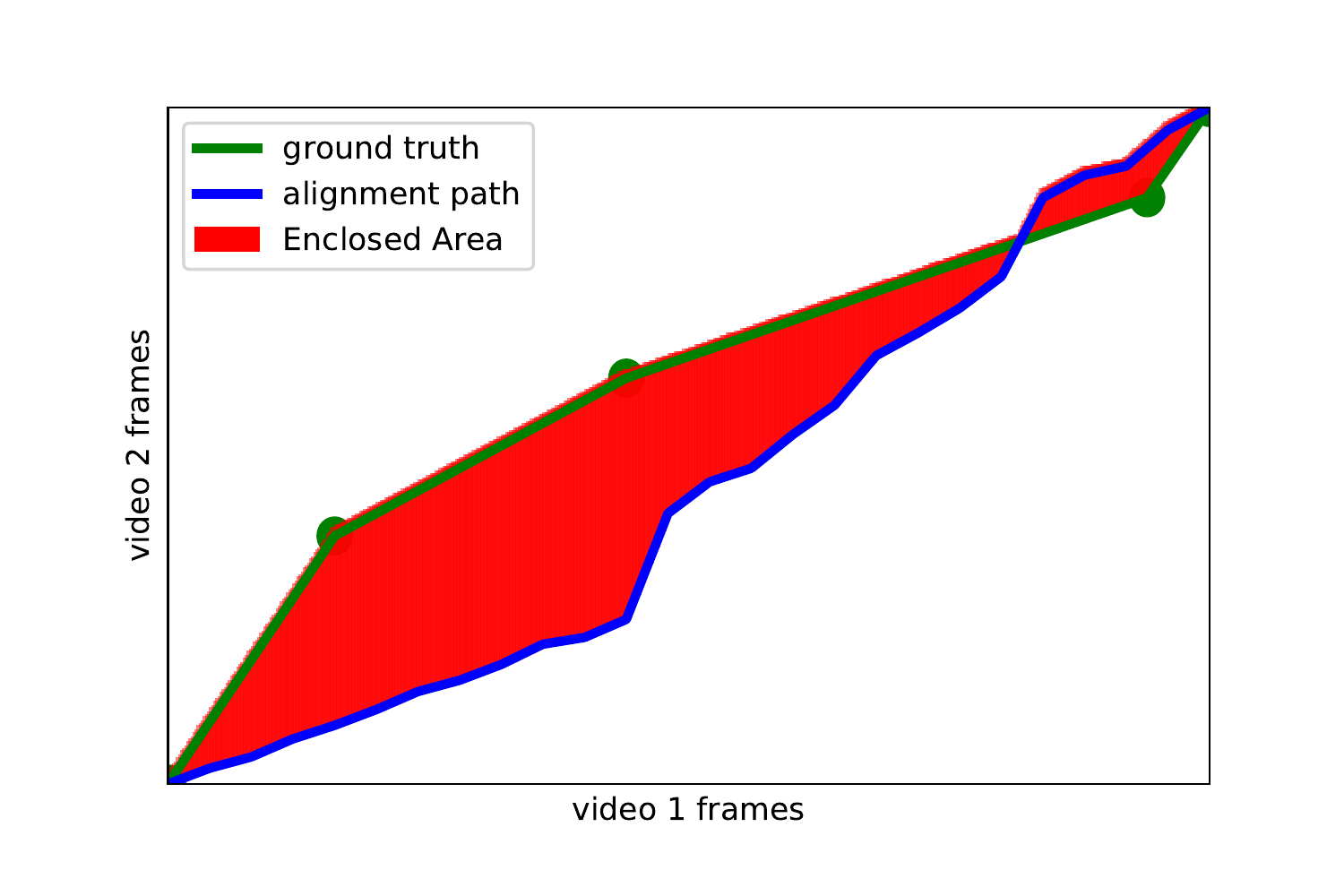}
  \caption{Enclosed area is the area between ground truth and predicted path. The EAE metric computes what fraction of the table's area is within the enclosed area.}
  \label{eae}
\end{figure}

It is almost impossible to manually align two videos frame-by-frame in order to use them as ground truth. Therefore, in the literature, each video is divided into a number of phases, and metrics such as phase classification and correct phase rate are used to evaluate the alignment output. However, the problem with such metrics is that there is no ground truth alignment between frames, and these metrics are only based on the number of individual frames that are aligned to a frame from the correct phase of the other video. In this section, a new metric for the alignment of two videos, or generally two sequences that consist of a number of phases, is introduced. During each phase, it is natural to suppose that the process is going forward linearly, which is a rather correct assumption for the Penn action dataset \cite{penn} and generally for human-action video datasets. By this assumption and knowing the boundaries of the phases in each video, a ground truth for the alignment can be obtained. We know some points on the ground truth path, and by the linearity assumption, the ground truth would be a piecewise-linear path going through those points(Figure \ref{eae}). The metric, which measures the deviation from the ground truth, then equals the area between the ground truth and alignment path divided by the whole area of the rectangle. 

\section{Datasets and Evaluation}
To evaluate and compare our approach with other studies, we employed three metrics using the Penn Action dataset \cite{penn} and a subset of UCF101 \cite{UCF}. This section introduces the datasets and the evaluation metrics used.

\subsection{Datasets}
The performance of our video alignment technique is evaluated on the Penn action dataset \cite{penn} and subset of UCF101 \cite{UCF} that called  UCF9 in this article. The complete list of all actions with the number of phases is given in Table \ref{tab:penn}.

\textbf{PennAction \cite{penn}}: This dataset comprises a diverse collection of actions performed by various individuals, with significant variations in execution. For comparative analysis, we utilized the phase labels provided by the authors of \cite{learning-by-aligning-videos-in-time} specifically for this dataset.

\textbf{UCF9}: In our study, a subset of the widely recognized UCF101 \cite{UCF} dataset is utilized, which contains 13,320 video clips across 101 action categories, all sourced from YouTube. The focus is placed on 9 specific actions from this dataset that are non-periodic and can be segmented into distinct phases. Each video corresponding to these actions is meticulously reviewed, and individual frames are annotated to accurately capture the different phases of the action.

\begin{table*}[t] 
    \begin{center}
    \begin{tabular}{|m{1cm} |m{3cm} |m{1cm} |m{11.5cm} |}
        \hline
        Dataset & Action & Number of Phases & List of Key Events  \\
        \hline
        \multirow{13}{4em}{Penn Action}
        & Baseball Pitch & 4 & Knee fully up, Arm fully stretched out, Ball release  \\
        & Baseball Swing & 3 & Bat swung back fully, Bat hits ball  \\
        & Bench Press & 2 & Bar fully down \\
        & Bowling & 3 & Ball swung fully back, Ball release\\
        & Clean and Jerk & 6 & Bar at hip, Fully squatting, Standing, Begin Thrusting, Beginning Balance\\
        & Golf Swing  & 3 & Stick swung fully back, Stick hits ball  \\
        & Jumping Jacks  & 4 & Hands at shoulder (going up), Hands above head, Hands at shoulders (going down)\\
        & Pullups  & 2 & Chin above bar\\
        & Pushups  & 2 & Head at floor\\
        & Situps  & 2 & Abs fully crunched\\
        & Squats  & 4 & Hips at knees (going down), Hips at floor, Hips at knee (going up) \\
        & Tennis Forehand & 3 & Racket swung fully back, Racket touches ball \\
        & Tennis Serve  & 4 & Ball released from hand, Racket swung fully back, Ball touches racket\\
        \hline
        \multirow{9}{4em}{UCF9}
        & Archery & 5 & Nock the Arrow, Draw and Anchor the Bow, aim, Release the String\\
        & Baseball Pitch & 4 & Knee fully up, Arm fully stretched out, Ball release \\
        & Clean and Jerk & 6 & Bar at hip, Fully squatting, Standing, Begin Thrusting, Beginning Balance\\
        & Field Hockey Penalty & 4 & Approach, Strike, Assessment\\
        & Golf Swing & 3 & Stick swung fully back, Stick hits ball \\
        & Hammer Throw  & 3 & Start turning body, Release Hammer\\
        & Javelin Throw  & 3 & Hand fully streched, Stop running to release\\
        & Long Jump  & 4 & Start jumping, Mid-air(Highest point), feet hit the ground\\
        & Pole Vault  & 6 & Preparation and Approach, Plant and Take-off, Swing Up, Turn and Push-off, Clearance and Landing\\
        \hline

    \end{tabular}
    \end{center}
    \caption{Number of phases for each activity in PennAction and UCF9 dataset.}
    \label{tab:penn}
\end{table*}

\subsection{Baselines}
For the experiments, in addition to using our proposed alignment method to align the videos, four other state-of-the-art methods are also implemented and used: GTA \cite{representation-learning-via-global-temporal-alignment}, TCC \cite{TCC}, TCN \cite{TCN}, and SAL \cite{SAL}. Moreover, a trivial baseline that aligns the frames of two videos linearly only based on their length is developed. This trivial method achieves outstanding results on the videos of this dataset because they have been trimmed, and extra/idle frames have been removed from them. An experiment is designed using synthesized videos that are closer to real-world videos to show the failure of the trivial method. The goal of this experiment is to show that the good performance of the trivial method on this dataset is by chance which is because of the special conditions of its videos. However, as it will be shown, our method is robust against various conditions.

\textbf{SAL \cite{SAL}}: This self-supervised method trains the Siamese network based on three-frame tuples and a classifier. The method randomly samples tuples from temporal windows with high motion and assigns positive or negative labels to them. This means that if the frames are in order, the tuple is considered positive, and if they are not in order, the tuple is considered negative. Additionally, the method trains the Classifier to predict whether the tuples are positive or negative.

\textbf{TCN \cite{TCN}}: In this work, a self-supervised method is introduced, involving a network trained on negative and positive frames. This method samples positives within a small window around anchors, while negatives are selected from distant timesteps within the same sequence. Additionally, the utilization of triplet loss in this work facilitates bringing anchors and positive frames closer to each other in the feature space compared to negative frames.

\textbf{TCC \cite{TCC}}: This self-supervised representation learning method trains a network using temporal cycle-consistency, which is a differentiable cycle-consistency loss that can find matched frames between two videos. This method produces per-frame embeddings for both videos and then uses Euclidian distance to align the frames of the second video to those of the first video.

\textbf{GTA \cite{representation-learning-via-global-temporal-alignment}}: In this work, a weakly-supervised representation learning method trains a network based on a loss that is composed of two parts. The first part consists of alignment losses, that are based on cumulative sum computations along optimal respective paths. These losses focus on achieving a consistent temporal alignment between frames of different videos. The second part of the loss involves global cycle consistency, utilized to ensure the stability of learned representations through cycle consistency.

\textbf{Trivial}: This method aligns two videos only based on their numbers of frames, using an assumption that in all videos, the process is going forward linearly. It means that if the first and second videos have $n$ and $m$ frames, respectively, then the $i-th$ frame of the first video is aligned to the $\frac{i}{n} \times j -th$ frame of the second video. Figure \ref{trivial-eae} shows EAE for trivial and our method on three pairs of videos.  

\subsection{Metrics}
In this work, three evaluation metrics are used: EAE, correct phase rate, and phase classification accuracy. EAE, which is our proposed metric, is explained in Section 2.3. This metric evaluates video synchronization task.

\textbf{Correct Phase Rate \cite{correct-phase-rate-1,correct-phase-rate-2}}: This metric finds the portion of the frames in the reference video that are aligned to any frame in the correct phase in the second video. This metric, which evaluates the video synchronization task, is calculated after frame alignment. 

\textbf{Phase classification accuracy}: This is the per frame phase classification accuracy on test data. To perform phase classification, an SVM is trained using the phase labels for each frame of the videos. Also, in this work, to ensure a fair comparison between the methods, care is taken to prevent overlap between the training data used for the feature extraction (in self-supervised methods) and the data used for SVM training. During the evaluation phase, a $10$-fold cross-validation is employed. In each iteration, one part of the test data is used as the training set for SVM, and the remaining data is reserved for testing purposes. This robust methodology is used for more accurate evaluation of the proposed method.

\subsection{Results}
The proposed method was evaluated on two tasks: phase classification and video synchronization.

In the phase classification task, frame classification is performed using the calculated features for each frame in conjunction with an SVM classifier. For the video synchronization task, the time series extracted from the videos are aligned using the DDTW method.

\subsubsection{Phase Classification}
Our final features are evaluated on the phase classification task and compared with the performance of SAL \cite{SAL}, TCN \cite{TCN}, TCC \cite{TCC} and GTA \cite{representation-learning-via-global-temporal-alignment} methods. In this setting, $10$-fold cross-validation is used on the test data. As shown in Table \ref{tab:Phase}, our method significantly outperforms the other methods in this task, demonstrating the effectiveness of the proposed features. 

It is important to note that video alignment is not performed during this experiment; the focus is solely on assessing the effectiveness of the extracted features for each frame. This experiment indicates that our features can effectively capture the crucial details necessary to distinguish between phases of the same actions. Additionally, since our approach does not require training data for the feature extraction process, it can be effectively used to label the phases of frames in new videos of the same action with high accuracy, even with only a few labeled videos.

\begin{table}[]
    \centering
    \begin{tabular}{|c|c|c|}
        \hline 
         Datasets & Method & Phase Classification  \\ \hline 
         \multirow{5}{4em}{Penn Action}
         & SAL & 47.69 \\
         & TCN & 58.2 \\
         & TCC & 69.6\\
         & GTA & \underline{76.3}\\
         & Ours & \textbf{80.92} \\ \hline
         \multirow{5}{4em}{UCF9}
         & SAL & 60.04 \\
         & TCN & 50.19 \\
         & TCC & 45.76\\
         & GTA & \underline{65.89}\\
         & Ours & \textbf{68.85} \\ \hline
    \end{tabular}
    \caption{Phase Classification Accuracy(\%)}
    \label{tab:Phase}
\end{table}

\subsubsection{Video Synchronization} 
In this experiment, each pair of videos is aligned, and the correct frame rate and Enclosed Area Error (EAE) are calculated for them. The results of video synchronization for the five methods are provided for each action in Table \ref{tab:main}. The results demonstrate that our final method, which combines feature vector extraction with Diagonalized Dynamic Time Warping (DDTW), significantly outperforms other state-of-the-art methods in video alignment.

\begin{table}[]
    \centering
    \begin{tabular}{|c|c|c|c|}
        \hline 
         Datasets & Method & EAE & Correct phase rate (\%) \\ \hline 
         \multirow{6}{4em}{Penn Action}
         & SAL & 0.15 & 68.3 \\
         & TCN & 0.107 & 74.7 \\
         & TCC & 0.125 & 71.7\\
         & GTA & 0.098 & 77.6\\
         & Trivial & \textbf{0.053} & \underline{81.8} \\
         & Ours & \underline{0.065} & \textbf{83.8} \\ \hline
         \multirow{6}{4em}{UCF9}
         & SAL & 0.203 & 54.18 \\
         & TCN & 0.146 & 61.56 \\
         & TCC & 0.168 & 59.95\\
         & GTA & 0.154 & 62.42\\
         & Trivial & \textbf{0.106} & \underline{63.24} \\
         & Ours & \underline{0.116} & \textbf{66.6} \\ \hline
    \end{tabular}
    \caption{Correct phase rate and EAE results on different activities}
    \label{tab:main}
\end{table}

Additionally, an experiment are conducted to show that the strong performance of the trivial method on the Penn and UCF9 datasets cannot be generalized to other datasets. The reason for the good performance of this method here is that the videos in the Penn action and UCF9 datasets are trimmed, i.e., idle frames are removed from the beginning and end of the videos. Also, there is no noticeable change in the speed of performing different phases of any action. In other words, the videos of the same action can be aligned with each other by linear expansion or shrinkage in the time domain. In this experiment, realistic modifications are applied to the videos to generate new ones for which the trivial method fails to align them successfully with the original videos.
 To this end, a "wait phase" is added at the beginning of all videos, by repeating the first three frames of the video; If the original video has $n$ frames, then $\frac{n}{2}$ frames are added at the beginning to generate a new video with $\frac{3n}{2}$ frames. This modification is realistic because it is natural to wait and concentrate before starting an exercise. As it is shown in table \ref{tab:trivial}, the trivial method fails to align modified videos to original ones effectively, and our method performs significantly better than trivial.
 
 \begin{table}[]
    \centering
    \begin{tabular}{|c|c|c|c|}
        \hline 
         Datasets & Method & EAE & Correct phase rate (\%) \\ \hline 
         \multirow{2}{4em}{Penn Action}
         & Trivial & 0.16 & 41.7 \\
         & Ours & \textbf{0.11} & \textbf{55.5} \\ \hline
         \multirow{2}{4em}{UCF9}
         & Trivial & 0.18 & 34.85 \\
         & Ours & \textbf{0.14} & \textbf{46.07} \\ \hline
    \end{tabular}
    \caption{The overall result of the experiment that shows the inefficiency of the trivial method on the other types of data.}
    \label{tab:trivial}
\end{table}

\section{Ablation Study}

In this section, we conduct a comprehensive ablation study to evaluate the effectiveness of different components of our proposed approach. The study is divided into two parts: (1) a comparison between Dynamic Time Warping (DTW) and Diagonalized Dynamic Time Warping (DDTW), and (2) an evaluation of the contributions of each component in our approach.

\subsection{Comparison of DTW and DDTW}

We first examine the impact of using DDTW instead of the standard DTW method for video alignment. Specifically, we compare the performance of state-of-the-art methods TCC \cite{TCC} and GTA \cite{representation-learning-via-global-temporal-alignment} on the UCF9 dataset using both DTW and DDTW. The results, presented in Table \ref{tab:dtw_vs_ddtw}, show that the application of DDTW significantly improves the correct phase rate, and reduces the EAE for both TCC and GTA. This improvement highlights the advantage of DDTW in capturing more accurate temporal correspondences between frames, thereby enhancing the overall performance of video alignment.

\begin{table}[]
    \centering
    \begin{tabular}{|>{\centering\arraybackslash}m{2.3cm} |>{\centering\arraybackslash}m{0.7cm} |>{\centering\arraybackslash}m{3.2cm} |}
        \hline 
         Method  & EAE & Correct phase rate (\%) \\ \hline 
         TCC + DTW & 0.168 & 59.95 \\
         TCC + DDTW & \textbf{0.148} & \textbf{61.87} \\
         \hline
         GTA + DTW & 0.154 & 62.42\\
         GTA + DDTW & \textbf{0.147} & \textbf{63.67}\\
         \hline
    \end{tabular}
    \caption{Comparison of DTW and DDTW on the UCF9 Dataset}
    \label{tab:dtw_vs_ddtw}
\end{table}

\subsection{Impact of Each Component in Our Approach}

We evaluate the contribution of each component in our approach by assessing the performance of the global and local features individually, and then in combination. Specifically, we compare three configurations on the UCF9 dataset: (1) VGG with kernel as global features, (2) box-pose features as local features, and (3) a combination of both global and local features. The results, provided in Table \ref{tab:component_study}, indicate that while each set of features performs reasonably well on its own, the combination of VGG kernel-based global features with box-pose local features yields the highest performance in terms of phase classification accuracy, correct phase rate, and EAE. This finding underscores the importance of integrating both global context and detailed local information to achieve superior video alignment and phase classification results.

\begin{table}[]
    \centering
    \begin{tabular}{|>{\centering\arraybackslash}m{2.6cm} |>{\centering\arraybackslash}m{1.8cm} |>{\centering\arraybackslash}m{0.7cm} |>{\centering\arraybackslash}m{1.7cm} |}
        \hline 
         Feature Configuration & Phase \newline Classification (\%) & EAE & Correct phase rate (\%) \\ \hline 
         VGG + Kernel \newline (Global Features) &  60.37 & 0.129 & 63.74 \\
         \hline
         Box-Pose \newline (Local Features) & 64.18 & 0.124 & 65.43 \\
         \hline
         Our & \textbf{68.85} & \textbf{0.116} & \textbf{66.6} \\
         \hline
    \end{tabular}
    \caption{Performance of Different Components on the UCF9 Dataset}
    \label{tab:component_study}
\end{table}

\subsection{Summary}

The results of the ablation study clearly demonstrate the advantages of using DDTW over DTW, and the significant performance boost achieved by combining global and local features. These findings validate the design choices in our approach and highlight the importance of each component in achieving state-of-the-art results.

\section{Conclusion}

This paper presents an unsupervised method for aligning two videos with the same action but different execution and appearance. In this method, a video is modeled as a multi-dimensional time series containing global and local, and static and dynamic features of its frames. A modified DTW method is also introduced for aligning time series, and a new metric is presented to compare the performance of time series alignment methods effectively. The results show that the proposed method provides significant performance improvement compared to the other methods and can be implemented for any action performed by one subject without any need for training any network. This work adds to the field of video alignment and has the potential to improve various video-related tasks such as action recognition, anomaly detection, and tracking.


{\small
\bibliographystyle{ieee_fullname}
\bibliography{Video-alignment}
}

\end{document}